\documentclass[conference]{IEEEtran}
\IEEEoverridecommandlockouts
\usepackage{cite}
\usepackage{amsmath,amssymb,amsfonts}
\usepackage{algorithmic}
\usepackage{graphicx}
\usepackage{textcomp}
\usepackage{threeparttable}
\usepackage{booktabs} 
\usepackage{multirow}
\usepackage{amsmath}
\usepackage{colortbl}
\usepackage{xcolor}
\usepackage{newfloat}
\usepackage{listings}
\usepackage{subcaption}
\def\BibTeX{{\rm B\kern-.05em{\sc i\kern-.025em b}\kern-.08em
    T\kern-.1667em\lower.7ex\hbox{E}\kern-.125emX}}
\begin{document}

\title{HT-GNN: Hyper-Temporal Graph Neural Network for Customer Lifetime Value Prediction in Baidu Ads
}

\author{
\IEEEauthorblockN{
Xiaohui Zhao\IEEEauthorrefmark{1}\IEEEauthorrefmark{4}, 
Xinjian Zhao\IEEEauthorrefmark{3}\IEEEauthorrefmark{4}, 
Jiahui Zhang\IEEEauthorrefmark{1},  
Guoyu Liu\IEEEauthorrefmark{1},
Houzhi Wang\IEEEauthorrefmark{1},
Shu Wu\IEEEauthorrefmark{2}
}
\vspace{-13pt}
\\
\IEEEauthorblockA{
\IEEEauthorrefmark{1}Baidu Inc.,
\IEEEauthorrefmark{2}Institute of Automation, Chinese Academy of Sciences,
\IEEEauthorrefmark{3}CUHK-Shenzhen
}
\{zhaoxiaohui03, zhangjiahui05, liuguoyu, wanghouzhi\}@baidu.com, xinjianzhao1@link.cuhk.edu.cn, shu.wu@nlpr.ia.ac.cn
\thanks{\IEEEauthorrefmark{4}Both authors contributed equally to this research.}
}

\maketitle

\begin{abstract}
Lifetime value (LTV) prediction is crucial for news feed advertising, enabling platforms to optimize bidding and budget allocation for long-term revenue growth. However, it faces two major challenges: (1) demographic-based targeting creates segment-specific LTV distributions with large value variations across user groups; and (2) dynamic marketing strategies generate irregular behavioral sequences where engagement patterns evolve rapidly.
We propose a Hyper-Temporal Graph Neural Network (HT-GNN), which jointly models demographic heterogeneity and temporal dynamics through three key components: (i) a hypergraph-supervised module capturing inter-segment relationships; (ii) a transformer-based temporal encoder with adaptive weighting; and (iii) a task-adaptive mixture-of-experts with dynamic prediction towers for 
multi-horizon LTV forecasting.
Experiments on \textit{Baidu Ads} with 15 million users demonstrate that HT-GNN consistently outperforms state-of-the-art methods across all metrics and prediction horizons.
\end{abstract}


\section{Introduction}
Predicting Customer Lifetime Value (LTV) has become a critical component of modern digital advertising ecosystems, directly shaping how platforms allocate resources, optimize user acquisition costs, and maximize long-term revenue \cite{badri2022beyond, schmittlein1987counting}. In the context of information feed advertising, accurate LTV prediction enables platforms to make data-driven decisions about ad placement strategies, user targeting, and marketing budget distribution \cite{he2024rankability, jain2002customer}. By predicting the total revenue a user will generate throughout their engagement with the platform, companies can dynamically adjust marketing strategies to achieve precise recommendations, thereby improving click-through rates, conversion rates, and ultimately generating substantial profits for advertisers.
\begin{figure}[h] \centering
\includegraphics[width=1.0\linewidth,keepaspectratio=true]{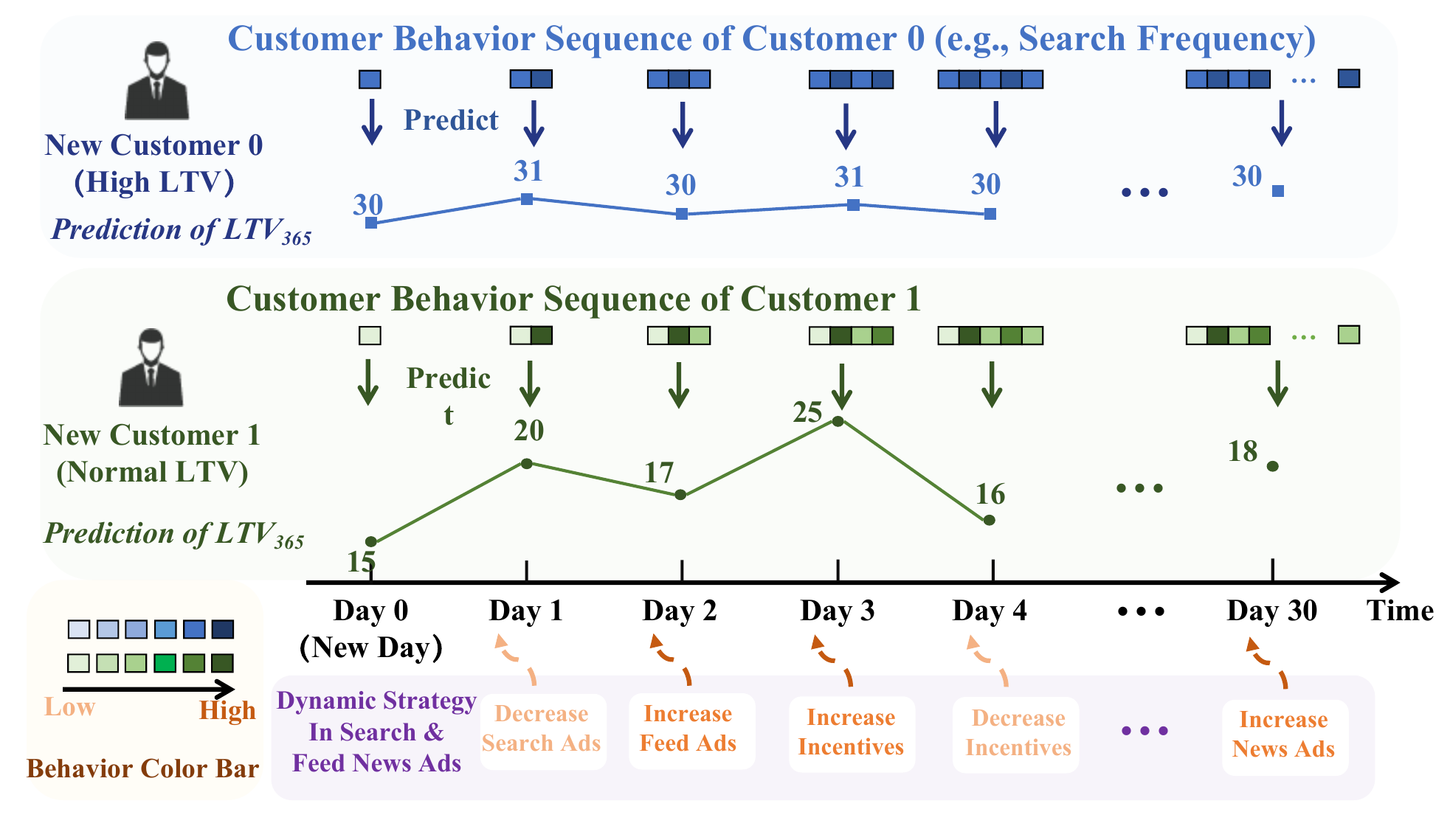}
    \caption{Illustration of uncertain behavior sequences under dynamic marketing strategies in advertising platforms}   
    \label{fig:LTV}
    \vspace{-10pt}
\end{figure}

LTV prediction has evolved through three generations of approaches. Early statistical methods modeled purchase behaviors as stochastic processes~\cite{fader2005rfm, norris1998markov}, while traditional machine learning approaches relied on handcrafted features, limiting their ability to capture temporal dependencies~\cite{lariviere2005predicting, singh2018customer}. Recent deep learning methods advanced the field through transformer-based sequence modeling~\cite{drachen2018or, yang2023feature, wang2024adsnet, xing2021learning}. However, these methods predominantly adopt an individual-user perspective and overlook structural relationships among user segments. Moreover, they cannot handle uncertain behavioral patterns induced by dynamic platform interventions in advertising environments. 

These overlooked aspects manifest as two fundamental challenges in advertising platforms. First, personalized ad delivery creates heterogeneous LTV distributions across user segments. Unlike traditional e-commerce, where user behaviors are relatively uniform, advertising platforms deliver customized content based on user profiles, leading to segment-specific engagement patterns and value distributions. Existing models, which assume homogeneous populations, fail to capture these inter-segment variations. Second, dynamic marketing strategies induce behavioral uncertainty. As illustrated in Fig.~\ref{fig:LTV}, platforms continuously adjust ad volume, pricing, and incentive mechanisms to optimize revenue, creating temporal patterns that differ fundamentally from organic user behaviors in content platforms \cite{yang2025use}. This strategic intervention makes behavioral sequences inherently noisy and difficult to predict.

We propose \textbf{HT-GNN} (Hyper-Temporal Graph Neural Network) to address these challenges through joint modeling of structural and temporal dimensions. The framework incorporates three key components: (1) hypergraph learning with JS-divergence supervision to capture inter-segment relationships while preserving segment-specific patterns, (2) transformer-based temporal encoding with adaptive weighting to handle irregular behavioral sequences, and (3) task-adaptive 
mixture-of-experts with dynamic towers and multi-loss optimization for robust multi-horizon prediction. By explicitly modeling both population structures and individual dynamics, HT-GNN provides a principled approach to LTV prediction in advertising environments.
Our main contributions include: 

\begin{itemize}
    \item We identify two fundamental challenges in advertising LTV prediction: heterogeneous user segments with distinct value distributions and behavioral uncertainty from dynamic platform interventions, validated through analysis of 15 million industrial users.
    \item We propose \textbf{HT-GNN}, a novel framework that integrates hypergraph learning for population modeling and adaptive temporal encoding for individual sequences, specifically designed for advertising platforms.
    \item Extensive experiments on real-world data from a major advertising platform demonstrate that HT-GNN significantly outperforms existing methods across all metrics and prediction horizons.
\end{itemize}
\section{RELATED WORK}

\subsection{LTV Prediction}

Traditional approaches employ statistical \cite{fader2005rfm, schmittlein1987counting} or classical machine learning methods \cite{lariviere2005predicting, singh2018customer}. While effective for aggregate predictions, these methods rely on handcrafted features and cannot capture temporal dependencies in user behaviors.

Deep learning has enabled more sophisticated modeling. ZILN \cite{wang2019deep} fits log-normal distributions via neural networks; TSUR \cite{xing2021learning} combines wavelet transforms with graph attention for sparse sequences; MDME \cite{li2022billion} employs mixture-of-experts across sub-distributions; expLTV \cite{zhang2023out} designs specialized detectors for high-spending users. However, these methods model users independently, overlooking group-level patterns that emerge from demographic targeting in advertising platforms. Moreover, they lack mechanisms to handle behavioral uncertainty induced by dynamic marketing interventions, a defining characteristic of advertising environments.
\subsection{Spatial-Temporal Graph Neural Networks}
Spatial-temporal GNNs have demonstrated effectiveness in modeling coupled spatial structures and temporal dynamics \cite{longa2023graph, jain2016structural, wang2024full}. Early architectures combined GNNs with RNNs \cite{li2017diffusion, seo2018structured, zhang2023doseformer}, while recent methods employ attention mechanisms for flexible integration \cite{zhang2018gaan, guo2019attention, vaswani2017attention, bai2021a3t, lu2024fed}. Applications span traffic prediction \cite{liu2023tap, peng2020spatial}, population flow modeling \cite{wu2020comprehensive, xu2021graph, von2015mobilitygraphs}, and recommendation systems \cite{xia2022hypergraph, xia2024ci, chen2023heterogeneous}. Advanced approaches \cite{yan20222, fan2021heterogeneous, wang2025hpst} tackle domain-specific challenges in UAV communication \cite{ren2023scheduling}, traffic flow \cite{bui2022spatial}, and social recommendations \cite{bai2020temporal}.
Despite success in these domains, spatial-temporal GNNs have not been adapted for advertising LTV prediction. The challenge lies in simultaneously handling demographic heterogeneity and behavioral uncertainty, characteristics absent in traffic or recommendation scenarios. HT-GNN addresses this gap through hypergraph learning that explicitly models inter-segment relationships while capturing individual temporal patterns shaped by dynamic marketing strategies.

\section{EMPIRICAL ANALYSIS}
While prior work has largely overlooked the joint influence of demographic structures and behavioral uncertainty, our empirical findings on real advertising data motivate a unified design.
In this section, we present empirical analyses that motivate our design choices for HT-GNN.

\begin{figure}[h]
    \centering
    \begin{subfigure}[t]{0.49\linewidth}
        \centering
        \includegraphics[width=\linewidth, keepaspectratio=true]{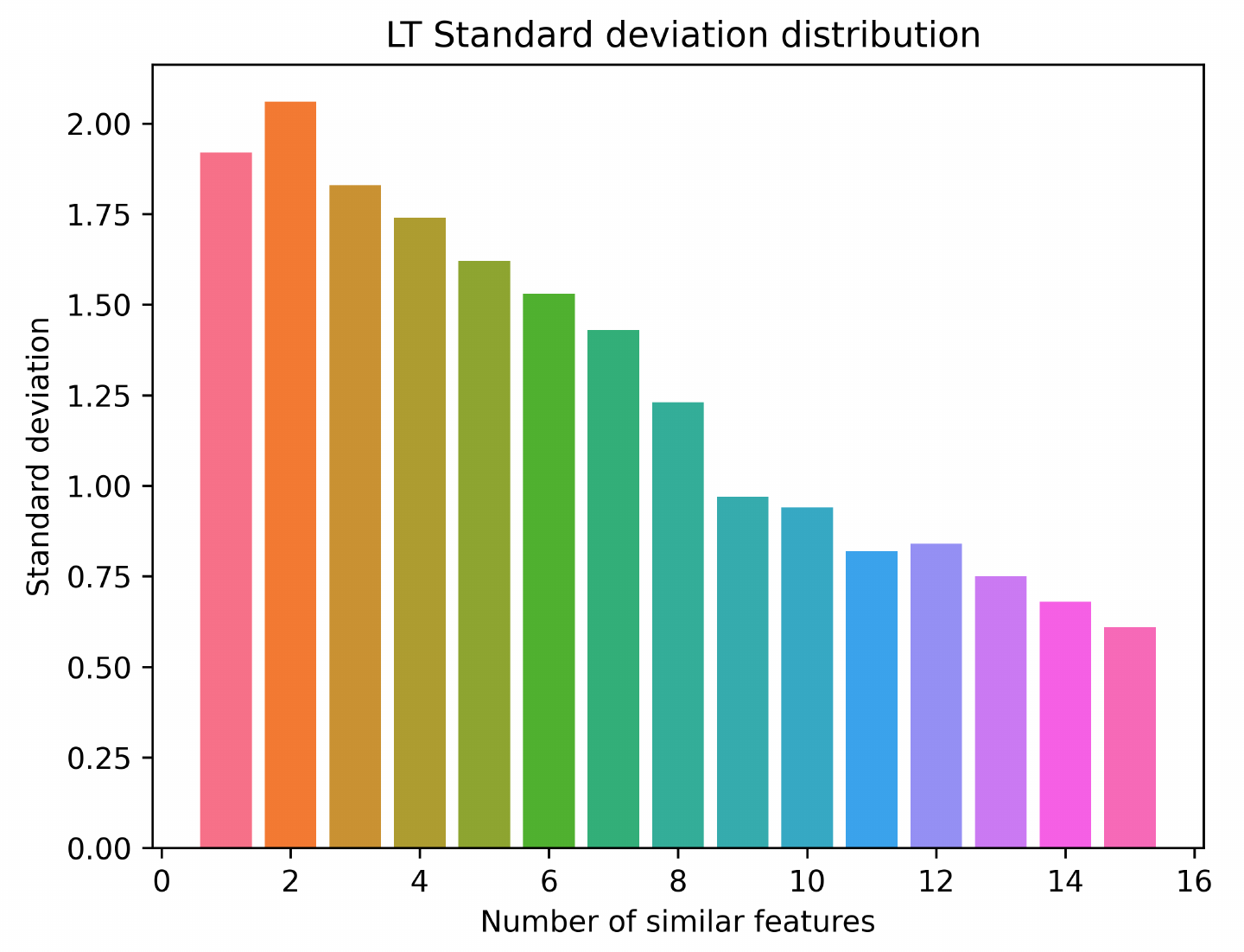}
        \caption{Lifetime (LT)}
        \label{fig:LTV differnt length}
    \end{subfigure}
    \hfill 
    \begin{subfigure}[t]{0.49\linewidth}
        \centering
        \includegraphics[width=\linewidth, keepaspectratio=true]{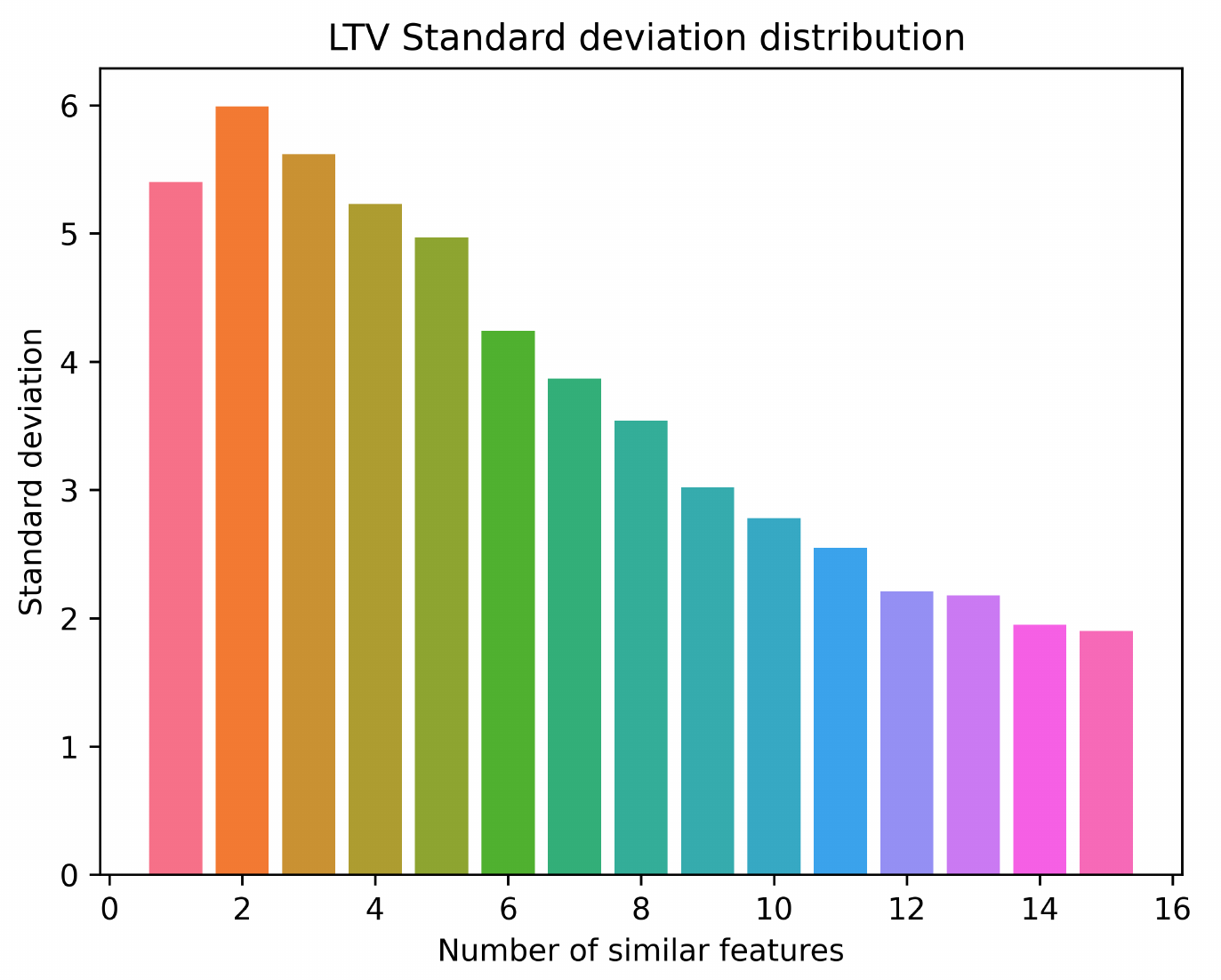}
        \caption{Lifetime Value (LTV)}
        \label{fig:LTV differnt length}
    \end{subfigure}
    \caption{Impact of Feature Similarity on LT and LTV Variance}
    \label{fig: differnt length}
\end{figure}

\noindent \textbf{Population Distribution.}
Early-stage LTV prediction faces a fundamental challenge: limited behavioral data. We observe that users with similar characteristics exhibit similar LTV patterns, which can be leveraged to improve predictions. To validate this, we analyzed how LTV variance changes when users share common features.
Fig. \ref{fig: differnt length} shows the standard deviation of LT and LTV values among users grouped by feature similarity. As the number of shared features increases (x-axis), the variance in both LT and LTV decreases significantly. For instance, users sharing 10+ features show 60-70\% lower variance compared to random groupings. This observation confirms that user groups exhibit stable regularities, motivating the use of hypergraph structures to explicitly encode such inter-segment relationships. 

\noindent \textbf{Challenges in LTV Distribution.}
Advertising LTV exhibits a heavily 
long-tailed distribution, with most users generating minimal revenue 
while few contribute disproportionately. Unlike e-commerce platforms with discrete purchase events, advertising platforms show continuous engagement patterns from search and feed ad interactions. This distribution poses two technical challenges: (1) standard MSE loss 
heavily penalizes errors on high-value users, leading to poor 
performance on the majority; (2) distributional assumptions from prior work, such as ZILN's log-normal assumption, do not hold. 
These observations motivate our multi-loss optimization strategy 
combining Huber loss for robustness and classification loss for 
zero-value user identification.

These empirical observations highlight the need for a model that simultaneously captures population-level regularities and temporal uncertainties, leading to our proposed HT-GNN framework.
\section{METHODOLOGY}
\subsection{Model Architecture}\label{AA}
\begin{figure*}[t] \centering
    \includegraphics[width=0.85\linewidth,keepaspectratio=true]{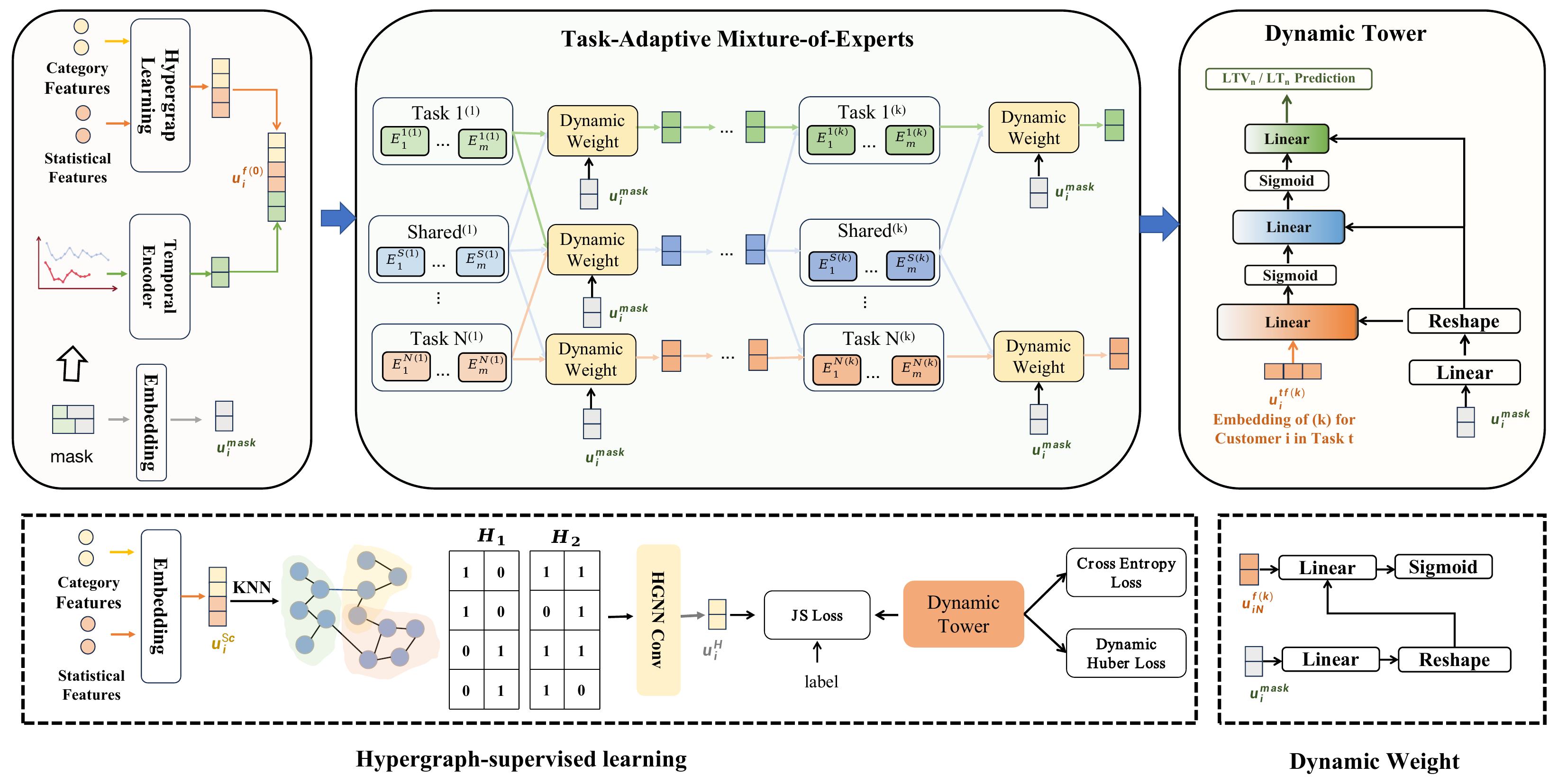}
    \caption{Hyper-Temporal Graph Neural Network  Modeling Framework}
    \label{fig:overview}
\end{figure*}
Fig. \ref{fig:overview} illustrates the overall architecture of HT-GNN, which consists of four main components: (1) Hypergraph Learning Module that captures high-order relationships among users, (2) Temporal Encoder that processes uncertain behavior sequences, (3) Task-Adaptive Mixture-of-Experts that handles heterogeneous user patterns, and (4) Dynamic Prediction Towers that generate robust predictions across multiple time horizons.

\subsection{Hypergraph-supervised learning} 
\noindent \textbf{Feature Representation.}
Given a user's categorical features and statistical features, we first transform them into a unified representation space. 
For categorical features, we apply one-hot encoding followed by embedding layers:
\begin{equation}
\mathbf{u}_{i}^c = \text{Embed}(\text{OneHot}(\mathbf{x}_i^c))
\end{equation}
Where the categorical feature representation of customer $i$ is denoted as $u_{i}^c$. 

For statistical features, we employ discretization to handle continuous values. Following Scott's rule, the bin width is:
\begin{equation}
d = \frac{3.5\sigma}{n^{1/3}}
\label{eq:bandwidth}
\end{equation}
where $\sigma$ represents the standard deviation, and $n$ represents the number of samples. The discretized features are then embedded as $\mathbf{u}_{i}^s$:
\begin{equation}
\mathbf{u}_{i}^s = \text{Embed}(\text{Discretize}(\mathbf{x}_i^s, d))
\end{equation}
The concatenated representation $\mathbf{u}_i^{sc} = [\mathbf{u}_{i}^c; \mathbf{u}_{i}^s]$ serves as the initial user representation.

\noindent \textbf{Hypergraph Construction.}
We construct a hypergraph $\mathcal{G} = (\mathcal{V}, \mathcal{E})$ where each node represents a user. Unlike simple graphs that only capture pairwise relationships, hyperedges can connect multiple nodes, naturally modeling group structures. For each node, we create a hyperedge connecting it to its $k$ nearest neighbors based on Euclidean distance in the feature space \cite{feng2019hypergraph}. The incidence matrix $\mathbf{H} \in \{0, 1\}^{|\mathcal{V}| \times |\mathcal{E}|}$ is defined as:
\begin{equation}
\mathbf{H}_{ve} = 
\begin{cases} 
1, & \text{if } v \in e \\
0, & \text{if } v \notin e
\end{cases}
\end{equation}
where \( v \in \mathcal{V} \) and \( e \in \mathcal{E} \). 
For a vertex $v \in \mathcal{V}$, its degree is defined as $d(v) = \sum_{e\in \mathcal{E}}\omega(e)\mathbf{H}_{ve}$. For an edge $e \in \mathcal{E}$, its degree is defined
as $\delta(e) = \sum_{v\in\mathcal{V}} \mathbf{H}_{ve}$. Further, $D_e$ and $D_v$ denote the diagonal matrices of the edge degrees and the vertex degrees.\\

\noindent \textbf{Hypergraph Convolution.} We perform message passing through hypergraph convolution:
\begin{equation}
{\mathbf{u}_{i}^{sc}}^{(l+1)} = \sigma\left({\mathbf{D}}_v^{-\frac{1}{2}} {\mathbf{H}} {\mathbf{W}} {\mathbf{D}}_e^{-1} {\mathbf{H}}^\top {\mathbf{D}}_v^{-\frac{1}{2}} {\mathbf{u}_{i}^{sc}}^{(l)} {\Theta}^{(l)}\right)
\end{equation}
where $\mathbf{W}$ is a diagonal matrix, and initialized as an identity matrix, which means equal weights for all hyperedges.

\noindent \textbf{Supervision via Jensen-Shannon Divergence.} 
A critical innovation is ensuring the learned representations preserve LTV-relevant structures. We introduce supervision through Jensen-Shannon (JS) divergence \cite{goodfellow2020generative} between feature space differences and LTV differences. 
Let $\mathbf{u}_{i}^{H}$ denote the output from the final hypergraph layer. We first compute the cosine similarity between user representations, yielding a similarity matrix $\mathbf{C}$ where $C_{ij} = \cos(\mathbf{u}_{i}^{H}, \mathbf{u}_{j}^{H})$. By applying softmax normalization to $1-\mathbf{C}$, we obtain a probability distribution matrix $\mathbf{M}$:
\begin{equation}
M_{ij} = \frac{\exp(1 - C_{ij})}{\sum_k \exp(1 - C_{ik})}
\end{equation}
where each row $M_i$ represents the normalized distribution of dissimilarities between user $i$ and all other users.

In order to quantify the dissimilarities among labels, we compute pairwise differences within each batch and subsequently transform them into a probability distribution via softmax normalization, yielding a probability distribution matrix $N$. Each row $N_i$ in the matrix represents the normalized distribution of label differences between user $i$ and all other users in the batch.Since our model employs uncertain sequence modeling, many users in the batch lack ground-truth labels (e.g., $LTV_{180}$ or $LTV_{365}$). For these samples, we use the model's predicted values as surrogate labels but apply a weighting factor $\frac{1}{e^{{\left| \mu_p - \mu_t \right|}*{\left| \sigma_p - \sigma_t \right|}}}$ to adjust their contribution during training, where $\mu_p,\sigma_p,\mu_t,\sigma_t$ represent the mean and variance of the predictive distribution and the mean and variance of the true (target) distribution, respectively.
Given a set of samples $\mathcal{T}$ with ground-truth labels and a set $\mathcal{F}$ without ground-truth labels, the Jensen-Shannon (JS) divergence for samples with true labels, denoted as $\mathcal{L}^{\text{JST}}$, can be written as:
\begin{equation}
\begin{aligned}
\mathcal{L}^{\text{JST}} = 
    &\frac{1}{2} \sum_{i \in \mathcal{T}} D_{\text{KL}}\left( M_i \parallel \frac{M_i + N_i}{2} \right) \\
\end{aligned}
\end{equation}
Similarly, the Jensen-Shannon (JS) divergence for samples using model predictions, denoted as $\mathcal{L}^{\text{JSP}}$, can be written as:
\begin{equation}
\begin{aligned}
\mathcal{L}^{\text{JSP}} = \frac{1}{2}\frac{1}{e^{{\left| \mu_p - \mu_t \right|}*{\left| \sigma_p - \sigma_t \right|}}} \sum_{j \in \mathcal{F}} D_{\text{KL}}\left( M_j \parallel \frac{M_j + N_j}{2} \right)
\end{aligned}
\end{equation}
Finally, the overall JS Loss is taken as the sum of the aforementioned two JS divergences:
\begin{equation}
\begin{aligned}
\mathcal{L}^{\text{JS}} = \mathcal{L}^{\text{JST}} + \mathcal{L}^{\text{JSP}}
\end{aligned}
\end{equation}
Where $D_{\text{KL}}(P\parallel Q) = \sum_{k} P(x_k) \log \left( \frac{P(x_k)}{Q(x_k)} \right)$ denotes the KL divergence between $P$ and $Q$. Since $\sum_{k} M_i(x_k)=\sum_{k} N_i(x_k) = 1$, the overall JS Loss can be calculated using the following formula:
\begin{equation}
\begin{aligned}
&\mathcal{L}^{\text{JS}} = 
    \frac{1}{2} \sum_{i \in \mathcal{T}} D_{\text{KL}} M_i(x_k) \log \left( \frac{M_i(x_k)}{M_i(x_k)+N_i(x_k)} \right) \\
    + &\frac{1}{2}\frac{1}{e^{{\left| \mu_p - \mu_t \right|}*{\left| \sigma_p - \sigma_t \right|}}} \sum_{j \in \mathcal{F}} D_{\text{KL}} N_j(x_k) \log \left( \frac{N_j(x_k)}{M_j(x_k)+N_j(x_k)} \right) 
\end{aligned}
\end{equation}

\subsection{Temporal encoder} 
For temporal features, we employ the Transformer encoder. We consider 35 distinct types of user behavior sequences, encompassing multi-dimensional attributes from users' interactions with in-feed advertisements.
For each type of sequence, we discretize the continuous values according to the bucketing method described in ~\eqref{eq:bandwidth}, then append a special token "\textit{[cls]}" at the beginning of the sequence. Subsequently, we apply varying degrees of masking to the sequence while recording its available length $s$. After embedding, positional encodings are added, and the resulting sequence is fed into the multi-head attention layer for computation:
\begin{equation}
o = softmax(\frac{QK^T+M}{\sqrt{d}} * V)
\end{equation}
Where $M$ represents our random-length mask matrix, $Q$, $K$, and $V$ are matrices after projection and  $d$ denotes the dimensionality after embedding. Next, we feed the multi-head attention layer's output o into a two-layer feedforward neural network (FFN) to obtain the holistic sequence feature representation. \\
We aggregate the feature vectors corresponding to all "\textit{[cls]}" tokens across sequences to obtain the ultimate user temporal representation $u_{i}^{se}$. Then, we utilize embedding layers to obtain the representation of the sequence-length mask, denoted as $u_{i}^{mask}$. Based on $u_{i}^{mask}$, we design a dynamic weighting mechanism. Finally, we extract the static categorical features, static statistical features, and dynamic uncertain behavior sequence features of customer $i$, denoted as $u_{i}^{f(0)}$.
\subsection{Task-Adaptive Mixture-of-Experts}

Different prediction horizons require different feature combinations—short-term predictions rely more on recent behaviors, while long-term predictions need stable user characteristics. We address this through a mixture-of-experts architecture that dynamically adapts to each task's requirements.

\noindent \textbf{Expert Architecture.} We design both task-specific experts $\{\mathbf{E}_j^t\}$ for each prediction horizon $t$ and shared experts $\{\mathbf{E}_j^s\}$ that capture common patterns. Each expert is a lightweight network:
\begin{equation}
E_j(\mathbf{x}) = \text{ReLU}(\text{BatchNorm}(\mathbf{W}_j\mathbf{x} + \mathbf{b}_j))
\end{equation}

The initial input to all experts is $\mathbf{u}_i^{(0)} = [\mathbf{u}_i^{sc}; \mathbf{u}_i^{se}]$, combining static features from hypergraph learning and temporal features from sequence encoding.

\noindent \textbf{Adaptive Gating Mechanism.} The key innovation is that expert selection depends on data availability. For task $t$ at layer $k$, we compute input-dependent weights based on the sequence mask:
\begin{equation}
\omega_{i}^{t(k)} = \sigma(\mathbf{W}_g^t \mathbf{u}_i^{mask} + \mathbf{b}_g^t)
\end{equation}

where the gating function considers both the previous layer's output and implicitly the sequence availability pattern.

\noindent \textbf{Task-Specific Expert Aggregation.} For each task $t$, we aggregate expert outputs:
\begin{equation}
\mathbf{E}_t = [E_1^t(\mathbf{u}_i^{t(k-1)}), ..., E_{n_t}^t(\mathbf{u}_i^{t(k-1)})]
\end{equation}
\begin{equation}
\mathbf{E}_s = [E_1^s(\mathbf{u}_i^{s(k-1)}), ..., E_{n_s}^s(\mathbf{u}_i^{s(k-1)})]
\end{equation}
\begin{equation}
\mathbf{u}_i^{t(k)} = \omega_i^{t(k)} \cdot [\mathbf{E}_t; \mathbf{E}_s]
\end{equation}

\noindent \textbf{Shared Expert Processing.} Shared experts aggregate information across all tasks to capture common patterns:
\begin{equation}
\mathbf{u}_i^{s(k)} = \omega_i^{s(k)} \cdot [\mathbf{E}_{T}; \mathbf{E}_s]
\end{equation}
where $\mathbf{E}_T$ contains outputs from all task-specific experts across all tasks. This design enables knowledge sharing while maintaining task-specific modeling capacity.
\subsection{Dynamic Tower}
For LT/LTV prediction in each time segment, we concurrently design a regression tower and a classification tower. The regression towers predict LTV$_t$/LT$_t$ values, while the classification towers identify whether LTV$_t$/LT$_t$ is zero. 
Within the task towers, we retain the dynamic weighting mechanism. For input $\mathbf{u}_{i}^{mask}$, we reshape it into three matrices, which are sequentially used as weights for three linear layers:
\begin{equation}
\mathbf{W}_1, \mathbf{W}_2, \mathbf{W}_3 = \text{Reshape}(\mathbf{W}_d \mathbf{u}_i^{mask} + \mathbf{b}_d)
\end{equation}
Additionally, we utilize the Sigmoid function between the linear layers. $W_3$ represents the output layer. For the classification tower, we add a sigmoid activation function at the end, while for the regression tower, we directly use the final output.
\subsection{Multi Loss Optimization}
LTV distributions in advertising are heavily long-tailed with many zero-valued users. We address this through a multi-objective optimization strategy.
For the classification task, we adopt the cross-entropy loss, which is specifically defined as follows:
\begin{equation}
\mathcal{L}^{\text{CE}} =  \sum_{i=1}^{n}c_{i}\log(\hat{c_i})
\end{equation}
where $c_i\in \{0,1\}$ is the true label, and $\hat{c_i}\in [0,1]$ is the prediction result from the classification tower.\\
To mitigate the interference from outliers and noise during the training process, for the regression task tower, we have designed a dynamic Huber Loss:
\begin{equation}
\mathcal{L}^{\text{H}}(y, \hat{y}) = 
\begin{cases} 
\frac{1}{2}(y - \hat{y})^2 & \text{if } |y - \hat{y}| \leq \delta, \\
\delta |y - \hat{y}| - \frac{1}{2}\delta^2 & \text{otherwise}.
\end{cases}
\end{equation}
Where $y$ represents the true LTV (Lifetime Value) for each subtask, $\hat{y}$ denotes the corresponding LTV predicted by the regression tower, and $\delta$ is the 95th percentile calculated from the error distribution $\{|y_i-\hat{y_i}|\}_{i=1}^{n}$ within each batch.\\
Finally, we define the multi-loss denoted as $\mathcal{L}$, as follows:
\begin{equation}
\mathcal{L} = \frac{1}{n} (\beta_{1}\sum_{i=1}^{N_r}  \mathcal{L}^{\text{JS}} + \beta_{2} \sum_{i=1}^{N_c}\mathcal{L}^{\text{CE}}\\
+ \sum_{i=1}^{N_r}\beta_{3}\mathcal{L}^{\text{H}}  )
\end{equation}
where $N_c$ represents the number of auxiliary classification towers, and $N_r$ represents the number of dynamic task towers, $\beta_1, \beta_2, \beta_3$ are trade-off parameters. 

\section{EXPERIMENT}
\subsection{Experimental Settings}
\noindent \textbf{Dataset.} We constructed a large-scale LTV prediction dataset from Baidu's information feed advertising platform, comprising 15 million user records collected between September 2023 and March 2025. The dataset includes user profiles, search and feed ad interactions, and 35 types of behavior sequences. We recorded actual LT/LTV values at 30, 180, and 365-day intervals. For evaluation, we split the data 9:1 for training/testing and stratified test samples based on their actual lifetime: users with 30-180 day lifetimes were evaluated on 30-day predictions, 181-365 day users on 180-day predictions, and \textgreater 365 day users on 365-day predictions.

\noindent \textbf{Evaluation Metrics.} We employ four complementary metrics: (1) \textbf{NRMSE} and \textbf{NMAE} for regression accuracy, where lower values indicate better performance; (2) \textbf{AUC} for classification performance in identifying active users; (3) \textbf{N-GINI} $\in [0,1]$ for ranking quality, measuring consistency between predicted and actual LTV rankings \cite{zhou2024cross, chen2025mini}. This comprehensive evaluation captures both value prediction accuracy and user segmentation effectiveness.
\subsection{Overall Performance}
\begin{table*}[t!]
\definecolor{bestcolor}{RGB}{255,230,230}
\definecolor{secondcolor}{RGB}{230,240,255}
\resizebox{\textwidth}{!}{%
\begin{threeparttable}
\renewcommand{\arraystretch}{1.05}


\begin{tabular}{@{}l|cccc|cccc|cccc@{}}
\toprule
\multirow{2}{*}{\textbf{Model}} & \multicolumn{4}{c|}{\textbf{LT\textsubscript{30}}} & \multicolumn{4}{c|}{\textbf{LT\textsubscript{180}}} & \multicolumn{4}{c}{\textbf{LT\textsubscript{365}}} \\
\cmidrule(lr){2-5} \cmidrule(lr){6-9} \cmidrule(lr){10-13}
 & NRMSE↓ & NMAE↓ & GINI↑ & AUC↑ & NRMSE↓ & NMAE↓ & GINI↑ & AUC↑ & NRMSE↓ & NMAE↓ & GINI↑ & AUC↑ \\
\midrule
XGBoost & 0.9845 & 0.5989 & 0.7384 & 0.7693 & 1.3227 & 0.9101 & 0.7437 & 0.7365 & 1.6678 & 0.8954 & 0.7101 & 0.6642 \\
ZILN & 0.7831 & 0.4386 & 0.8002 & 0.8120 & 1.0604 & 0.8329 & 0.7940 & 0.7765 & 1.3865 & 0.8003 & 0.7428 & 0.6883 \\
expLTV & 0.6983 & 0.3123 & \cellcolor{secondcolor}0.9133 & \cellcolor{secondcolor}0.8891 & 0.9997 & 0.8217 & \cellcolor{secondcolor}0.8268 & 0.8002 & 1.1482 & 0.7829 & \cellcolor{secondcolor}0.7995 & 0.7168 \\
HST-GT & 0.5657 & 0.2431 & 0.9067 & 0.8787 & 0.9365 & 0.7904 & 0.8118 & \cellcolor{secondcolor}0.8095 & 1.0688 & 0.6288 & 0.7483 & \cellcolor{secondcolor}0.7236 \\
MDME & \cellcolor{secondcolor}0.4712 & \cellcolor{secondcolor}0.1891 & 0.8993 & 0.8762 & \cellcolor{secondcolor}0.8996 & \cellcolor{secondcolor}0.6557 & 0.8223 & 0.7903 & \cellcolor{secondcolor}0.9987 & \cellcolor{secondcolor}0.6283 & 0.7899 & 0.7095 \\
\textbf{HT-GNN} & \cellcolor{bestcolor}\textbf{0.1109} & \cellcolor{bestcolor}\textbf{0.0460} & \cellcolor{bestcolor}\textbf{0.9987} & \cellcolor{bestcolor}\textbf{0.9979} & \cellcolor{bestcolor}\textbf{0.7128} & \cellcolor{bestcolor}\textbf{0.4168} & \cellcolor{bestcolor}\textbf{0.8909} & \cellcolor{bestcolor}\textbf{0.8463} & \cellcolor{bestcolor}\textbf{0.8083} & \cellcolor{bestcolor}\textbf{0.4814} & \cellcolor{bestcolor}\textbf{0.8466} & \cellcolor{bestcolor}\textbf{0.7292} \\
\midrule
\multirow{2}{*}{\textbf{Model}} & \multicolumn{4}{c|}{\textbf{LTV\textsubscript{30}}} & \multicolumn{4}{c|}{\textbf{LTV\textsubscript{180}}} & \multicolumn{4}{c}{\textbf{LTV\textsubscript{365}}} \\
\cmidrule(lr){2-5} \cmidrule(lr){6-9} \cmidrule(lr){10-13}
 & NRMSE↓ & NMAE↓ & GINI↑ & AUC↑ & NRMSE↓ & NMAE↓ & GINI↑ & AUC↑ & NRMSE↓ & NMAE↓ & GINI↑ & AUC↑ \\
\midrule
XGBoost & 1.1709 & 0.7166 & 0.7662 & 0.7867 & 3.1892 & 1.0988 & \cellcolor{secondcolor}0.7363 & 0.7567 & 3.1224 & 1.0565 & 0.6664 & 0.6867 \\
ZILN & 1.0526 & 0.6527 & 0.7034 & 0.7883 & 2.9989 & 0.9996 & 0.6894 & 0.7018 & 2.9856 & 0.9796 & 0.6543 & 0.6258 \\
expLTV & 0.9915 & 0.5415 & 0.8462 & \cellcolor{secondcolor}0.8875 & 2.9079 & 0.9344 & \cellcolor{bestcolor}0.7491 & \cellcolor{secondcolor}0.7572 & 2.8986 & 0.9366 & \cellcolor{secondcolor}0.7105 & \cellcolor{bestcolor}0.7288 \\
HST-GT & 0.9602 & 0.4837 & \cellcolor{secondcolor}0.8977 & 0.8499 & 2.8886 & 0.8933 & 0.7264 & 0.7102 & 2.5969 & 0.9269 & 0.7101 & 0.7107 \\
MDME & \cellcolor{secondcolor}0.8786 & \cellcolor{secondcolor}0.3665 & 0.8169 & 0.8740 & \cellcolor{secondcolor}2.6528 & \cellcolor{secondcolor}0.8492 & 0.7292 & 0.7488 & \cellcolor{secondcolor}2.3234 & \cellcolor{secondcolor}0.8696 & 0.7001 & \cellcolor{secondcolor}0.7267 \\
\textbf{HT-GNN} & \cellcolor{bestcolor}\textbf{0.7476} & \cellcolor{bestcolor}\textbf{0.1969} & \cellcolor{bestcolor}\textbf{0.9955} & \cellcolor{bestcolor}\textbf{0.9361} & \cellcolor{bestcolor}\textbf{2.1697} & \cellcolor{bestcolor}\textbf{0.7551} & \textbf{0.6843} & \cellcolor{bestcolor}\textbf{0.8228} & \cellcolor{bestcolor}\textbf{1.9581} & \cellcolor{bestcolor}\textbf{0.7769} & \cellcolor{bestcolor}\textbf{0.7318} & \textbf{0.7254} \\
\bottomrule
\end{tabular}
\end{threeparttable}
}
\caption{Performance comparison with different approaches. Best results are highlighted in \colorbox{bestcolor}{red}, second-best in \colorbox{secondcolor}{blue}. ↓ indicates lower is better, ↑ indicates higher is better.}
\label{tb:baseline}
\end{table*}

\noindent \textbf{Baselines.}   
We compare HT-GNN with five representative methods spanning traditional machine learning to recent deep learning approaches. \textbf{XGBoost} \cite{drachen2018or} employs a two-stage process that first classifies users as paying or non-paying, then predicts their monetary value. \textbf{ZILN} \cite{wang2019deep} assumes LTV follows a log-normal distribution and uses neural networks to fit distribution parameters. \textbf{MDME} \cite{li2022billion} divides the LTV distribution into multiple sub-distributions and buckets for more granular modeling. \textbf{expLTV} \cite{zhang2023out} specifically targets gaming scenarios by separately modeling high-spending and low-spending users. HST-GT \cite{zhao2023hst}, originally designed for e-commerce delivery time estimation, applies heterogeneous spatial-temporal graphs to capture complex relationships.

Table \ref{tb:baseline} shows that HT-GNN consistently outperforms all baselines across all prediction horizons. Notably, HT-GNN achieves over 75\% relative improvement on short-term (LT30/LTV30) tasks compared to the second-best model (MDME). The improvements stem from three key aspects: (1) our hypergraph learning captures group-level patterns missed by individual modeling approaches, (2) the adaptive multi-task framework handles heterogeneous user behaviors better than fixed architectures, and (3) our multi-loss strategy is more robust to long-tailed distributions than methods with strict distributional assumptions such as ZILN and expLTV. 
\subsection{Ablation Study }
We conducted ablation studies to examine the effects of individual components within the HT-GNN framework. Three method variants were compared in the experiments: (1) w/o HG: using only the Temporal encoder while ignoring structural connections between users; (2) w/o DW: abandoning the Task-Adaptive Mixture-of-Experts and employing only the classic multi-task learning paradigm; (3) w/o DT: disregarding the dynamic towers and utilizing only the simplest linear layer or sigmoid classification layer. Table II presents a comparison of the three variant models of HT-GNN on two subtasks, namely $LT_{30}$ and $LTV_{30}$.

\begin{table}[t]
\centering
\begin{tabular}{lcccc} 
\toprule
\textbf{Model} & \multicolumn{2}{c}{\textbf{LT\textsubscript{30}}} & \multicolumn{2}{c}{\textbf{LTV\textsubscript{30}}} \\ 
\cmidrule(lr){2-3} \cmidrule(lr){4-5} 
& \textbf{NRMSE} & \textbf{NMAE} & \textbf{NRMSE} & \textbf{NMAE} \\ 
\midrule
\textit{w/o HG} & 0.4866 & 0.2333 & 1.4446 & 0.4730\\ 
\textit{w/o DW} & 0.2441 & 0.1523 & 1.2245 & 0.4589\\ 
\textit{w/o DT} & 0.2238 & 0.1399 & 1.3087 & 0.4868\\ 
\textbf{HT-GNN} & \textbf{0.1109} & \textbf{0.0460} & \textbf{0.7476} & \textbf{0.1969} \\ 
\bottomrule
\end{tabular}
\caption{Ablation Study of Hypergraph Learning, Dynamic Weight Mechanism and Dynamic Tower Model}
\label{tab:ablation_study} 
\end{table}
(i) When we do not employ hypergraph mining to extract population structure information, w/o HG performs worse than HT-GNN, demonstrating that extracting implicit associations among user populations is essential for boosting LTV prediction performance.

(ii) When we do not consider dynamic weights for uncertain sequences, w/o DW performs worse than HT-GNN, indicating that uncertainty information between customer behavior sequences allows the model to capture distinct user profiles for better personalization.

(iii) When dynamic parameters are removed, w/o DT shows performance deterioration, indicating the positive effect of fusing customer embeddings and uncertain sequences.


\subsection{Multi-Loss Impact Analysis }
In this section, we conduct a comparative analysis of different loss functions to validate the effectiveness of our loss selection strategy. We compares three loss selection strategies: (1) optimizing the model solely with MSE Loss, (2) optimizing the model solely with Huber Loss, and (3) jointly optimizing the model with both Huber Loss and JS Loss.
 Shown in Table~\ref{tab:loss_study}, we present the experimental results of three loss selection approaches of HT-GNN on two subtasks.

\begin{table}[t]
\centering
\begin{tabular}{lcccc} 
\toprule
\textbf{Loss} & \multicolumn{2}{c}{\textbf{LT\textsubscript{30}}}& \multicolumn{2}{c}{\textbf{LTV\textsubscript{30}}} \\ 
\cmidrule(lr){2-3} \cmidrule(lr){4-5} 
& \textbf{NRMSE} & \textbf{NMAE} & \textbf{NRMSE} & \textbf{NMAE} \\ 
\midrule
\textbf{MSE Loss} & 0.2556 & 0.1309 & 2.5003 & 0.7508\\ 
\textbf{Huber Loss} & 0.2203 & 0.1448 & 1.3596 & 0.4126\\ 
\textbf{Multi-Loss} & \textbf{0.1109} & \textbf{0.0460} & \textbf{0.7476} & \textbf{0.1969} \\ 
\bottomrule
\end{tabular}
\caption{Loss Selection of MSE, Huber and Multi-Loss}\label{tab:loss_study} 
\end{table}
The results in Table~\ref{tab:loss_study} indicate that when only the MSE Loss is employed, the LT prediction exhibits superior performance, likely because LT values are integers ranging from 1 to 30, resulting in relatively stable mean values. Conversely, when solely utilizing the Huber Loss, the LTV prediction achieves better results, presumably due to the heightened sensitivity of MSE to noise and outliers within the LTV distribution. Notably, the model incorporating both dynamic Huber Loss and JS Loss demonstrates a significant performance enhancement, with approximately $50\%$ improvements observed in both NRMSE and NMAE metrics. This substantiates the effectiveness of our loss selection strategy.


\subsection{Comparison of GNN Architectures}
Table~\ref{tab:GNN} compares HT-GNN against classical graph learning algorithms including GCN \cite{kipf2016semi}, GraphSAGE \cite{hamilton2017inductive}, and GAT \cite{velivckovic2017graph}. Hypergraph learning outperforms other graph learning algorithms. Traditional graph structures can only depict pairwise relationships between two nodes (i.e., first-order relationships), yet they fall short in addressing higher-order correlations, thus constraining their modeling capabilities. Hypergraphs are capable of modeling more flexible and intricate relationships, exhibiting stronger expressive power and consequently demonstrating superior performance in complex user interactions.


\begin{table}[t]
\centering
\begin{tabular}{lcccc} 
\toprule
\textbf{Model} & \multicolumn{2}{c}{\textbf{LT\textsubscript{30}}} & \multicolumn{2}{c}{\textbf{LTV\textsubscript{30}}}\\ 
\cmidrule(lr){2-3} \cmidrule(lr){4-5} 
& \textbf{NRMSE} & \textbf{NMAE} & \textbf{NRMSE} & \textbf{NMAE} \\ 
\midrule
\textbf{GCN} & 0.2168 & 0.1203 & 0.9596 & 0.2906\\ 
\textbf{GraphSAGE} & 0.1936 & 0.1189 & 0.9577 & 0.2799\\ 
\textbf{GAT} & 0.1923 & 0.1148 & 1.0502 & 0.3164\\ 
\textbf{HT-GNN} & \textbf{0.1109} & \textbf{0.0460} & \textbf{0.7476} & \textbf{0.1969} \\ 
\bottomrule
\end{tabular}
\caption{Comparison of GNN Performance}\label{tab:GNN} 
\vspace{-10pt} 
\end{table}




\section{Conclusion}
We address the LTV prediction challenge in advertising platforms through HT-GNN, a framework that integrates hypergraph learning and adaptive temporal modeling. By explicitly capturing inter-segment relationships and individual behavioral dynamics, HT-GNN overcomes limitations of traditional user-independent approaches. Experiments on Baidu's advertising data demonstrate substantial improvements in prediction accuracy, particularly for users exhibiting complex behavioral patterns under dynamic marketing strategies. This work establishes a foundation for incorporating structural and temporal reasoning into value prediction systems, with potential applications beyond advertising to domains featuring heterogeneous user populations and evolving behavioral contexts.

\bibliographystyle{IEEEtran}
\bibliography{ICDE2026}
\end{document}